\documentclass[letterpaper]{article} %
\usepackage{aaai24}  %
\usepackage{times}  %
\usepackage{helvet}  %
\usepackage{courier}  %
\usepackage[hyphens]{url}  %
\usepackage{graphicx} %
\usepackage{natbib}  %
\usepackage{caption} %
\usepackage{algorithm}
\usepackage{algorithmic}

\usepackage{newfloat}
\usepackage{listings}
\DeclareCaptionStyle{ruled}{labelfont=normalfont,labelsep=colon,strut=off} %
\floatstyle{ruled}
\newfloat{listing}{tb}{lst}{}
\floatname{listing}{Listing}
\usepackage[utf8]{inputenc} %
\usepackage[T1]{fontenc}    %
\usepackage{booktabs}       %
\usepackage{amsfonts}       %
\usepackage{nicefrac}       %
\usepackage{microtype}      %
\usepackage{xcolor}         %
\usepackage{graphicx}
\usepackage{makecell}
\usepackage{multirow}
\usepackage{times}
\usepackage{epsfig}
\usepackage{amsmath}
\usepackage{amssymb}
\usepackage{placeins}
\usepackage{comment}

\usepackage{xcolor,colortbl}
\usepackage{rotating}
\usepackage{subcaption}

\newcommand{\I}{\mathcal{I}}
\newcommand{\F}{\mathcal{F}}
\newcommand{\Faug}{\mathcal{F}'}
\newcommand{\SQ}{S}
\newcommand{\SR}{S}

\renewcommand{\P}{\mathbf{P}}
\newcommand{\R}{\mathbf{R}}
\renewcommand{\t}{\mathbf{t}}
\newcommand{\Lregr}{\mathcal{L}_{\text{regr}}}

\newcommand{\Lfinal}{\mathcal{L}_{\text{final}}}
\newcommand{\y}{\mathbf{y}}

\newcommand{\yh}{\hat{\mathbf{y}}}

\newcommand{\fattn}{f_{\text{attn}}}

\newcommand{\ie}{\textit{i}.\textit{e}. }
\newcommand{\wrt}{\textit{w}.\textit{r}.\textit{t}. }
\newcommand{\PAR}[1]{\vspace{1mm}\noindent{\bf{#1}}}
\newcommand{\eg}{\textit{e}.\textit{g}. }

\newlength{\SCRfigwidth}
\newlength{\SCRfigheight}
\newcommand{\SCRraisedgraphics}[1]{\raisebox{-.5\height}[\dimexpr0.5\height+2pt]{\includegraphics[width=\SCRfigwidth, height=\SCRfigheight, keepaspectratio]{#1}}}
\newenvironment{SCRfigtabular}
{
    \setlength{\SCRfigwidth}{0.16\linewidth}
    \setlength{\SCRfigheight}{\SCRfigwidth}
    \setlength{\tabcolsep}{3pt}
    \bf{}
    \small{}
    \begin{tabular}{ccc|c|cc}
    \multicolumn{3}{c|}{Reference images with proxy shape}
    & Query image 
    & Predicted coordinates
    & Confidence \\
}
{
    \end{tabular}
}
\newcommand{\SCRfigline}[1]
{
    \SCRraisedgraphics{#1/ref1_overlayed.png} &
    \SCRraisedgraphics{#1/ref2_overlayed.png} &
    \SCRraisedgraphics{#1/ref3_overlayed.png} &
    \SCRraisedgraphics{#1/q_pred_bbox.png} &
    \SCRraisedgraphics{#1/q_pred3d.jpg} &
    \SCRraisedgraphics{#1/q_conf.jpg}
}

\title{MFOS: Model-Free \& One-Shot Object Pose Estimation}
\author{
    JongMin Lee\textsuperscript{\rm 1}
    Yohann Cabon\textsuperscript{\rm 3}
    Romain Brégier\textsuperscript{\rm 3}
    Sungjoo Yoo\textsuperscript{\rm 2}
    Jerome Revaud\textsuperscript{\rm 3}
}
\affiliations{
    \textsuperscript{\rm 1}\textsuperscript{\rm 2}Seoul National University \quad\quad\quad\quad\quad\quad\quad\quad\quad \textsuperscript{\rm 3}Naver Labs Europe\\
    \textsuperscript{\rm 1}sdrjseka96@naver.com \quad
    \textsuperscript{\rm 2}sungjoo.yoo@gmail.com \quad \textsuperscript{\rm 3}\texttt{firstname.lastname@naverlabs.com}
}

\usepackage{bibentry}

\begin{document}

\maketitle

\begin{abstract}
Existing learning-based methods for object pose estimation in RGB images are mostly model-specific or category based.
They lack the capability to generalize to new object categories at test time, hence severely hindering their practicability and scalability.
Notably, recent attempts have been made to solve this issue, but they still require accurate 3D data of the object surface at both train and test time. %
In this paper, we introduce a novel approach that can estimate in a single forward pass the pose of objects never seen during training, given minimum input. %
In contrast to existing state-of-the-art approaches, which rely on task-specific modules, our proposed model is entirely based on a transformer architecture, %
which can benefit from recently proposed 3D-geometry general pretraining.
We conduct extensive experiments and report state-of-the-art one-shot performance on the challenging  LINEMOD benchmark.
Finally, extensive ablations allow us to determine good practices with this relatively new type of architecture in the field. 

\end{abstract}

\section{Introduction}
Being able to estimate the pose of objects in an image is a mandatory requirement for any tasks involving some kind of interactions with objects. 
The past decade has seen a surge of research in 3D vision, with potential applications ranging from robotics~\cite{robot_pose_1, robot_pose_2} to VR/AR~\cite{ar_pose1, ar_pose2}. 
These applications require pose estimators that are accurate, robust and scalable.
In this context, we tackle the problem of object pose estimation from a single image, \ie we aim at extracting the 6D pose of a target object relatively to the camera. %

Object pose estimation is a long-studied research topic. %
Earlier approaches were holistic~\cite{holistic1, holistic2, holistic3, holistic4, holistic5}, based on sliding-window template-based matching~\cite{sliding1, sliding2}
or utilized local feature matching~\cite{local_feature1, local_feature2, local_feature3}.
In all cases, these methods were heavily handcrafted and yielded unsatisfactory results regarding robustness and accuracy.
With the advent of deep learning, a new training-based paradigm emerged for object pose estimation~\cite{posecnn, cdpn, GDRnet, pix2pose},
the idea of letting a deep network end-to-end predict the pose of an object from an image, given sufficient training data (images of the same object in various poses).
While significantly improving in robustness and accuracy, these methods have the disadvantage of being \emph{model-specific}: they can only cope with objects seen during training.

While some works have broadened the model scope to object categories rather than object instances~\cite{nocs,shape_category, category_scale_shape, category_size_space, category_synthesis}, the trained model is still only suitable for objects or categories seen during training. 
To remedy this shortcoming, recent learning-based methods that can generalize to unseen objects, denoted as "one-shot", have been proposed.
In practice, however, they rely on 3D models~\cite{ove6d, osop}, require video sequences~\cite{bundletrack} or additional depth maps~\cite{fs6d} at test time. 
All in all, these requirements severely hinder their practicality and scalability.

\begin{figure*}
    \centering
    \includegraphics[trim=0 280 110 0,clip,width=\linewidth]{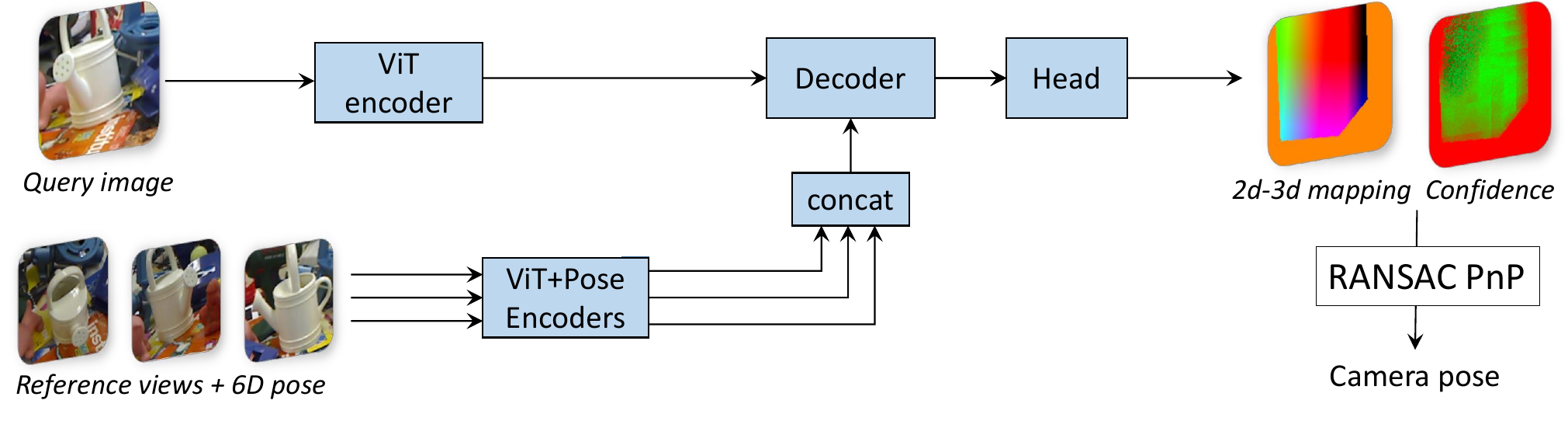} \\[-0.3cm]
    \caption{\textbf{Overview of the method}. 
    Our model takes as input a query image and a set of $K$ reference views of the same object seen under different viewpoints (annotated with pose information).
    We use a vision transformer (ViT) to first encode all images.
    For reference images, their corresponding object pose is jointly encoded with the image.
    Then, a transformer decoder jointly processes features from the query and reference images.
    Finally, a prediction head outputs a dense 2D-3D mapping and a corresponding confidence map, from which we can recover the query object pose by solving a PnP problem.
    }
    \label{fig:overview}
\end{figure*}

In this paper, we propose a novel approach to address the limitations of previous methods for object pose estimation. 
As illustrated in Figure~\ref{fig:overview}, our method can estimate the pose of a target object from a single image, denoted as \emph{query} image in the following.
To estimate the target object pose, the only required inputs at inference time are a rough estimate of the object size and a small collection of \emph{reference} images of the target object with known poses. %
These inputs can be obtained via scalable and straightforward methods, \eg fiducial markers (AprilTags) \cite{apriltag} or SfM~\cite{colmap}.  %
Similar to previous work, our model outputs a dense 2D-3D mapping from which the object pose can be obtained straightforwardly~\cite{dpod, cdpn, pix2pose}. %

Our approach is entirely implemented using Vision Transformer (ViT) blocks~\cite{vit}. %
Doing so enables us to leverage a powerful pretraining technique specifically tailored to 3D vision that can embed strong geometric priors in the network.
Specifically, we initialize our network from an off-the-shelf model pretrained using Cross-View Completion (CroCo)~\cite{crocostereo}.
We show that this pretraining considerably boost the generalization capabilities of our method, making it possible to estimate the pose of target objects unseen during training.
Inspired by BB8~\cite{bb8}, we simply yet effectively encode the object pose with a \emph{proxy shape} positioned and scaled according to the object's pose and dimensions, respectively. 
We show that using rectangular cuboid as proxy shape works well in practice and allows us to deal with objects of unknown shape at test time.
Our overall architecture is a generalization of the CroCo architecture~\cite{croco} to multiple reference images (instead of just one in CroCo).
It is computationally efficient at both training and test time, and it requires a single forward pass. %

To ensure robust generalization of our model, we train it on a diverse set of object-centric data, including the BOP dataset~\cite{bop_dataset}, OnePose~\cite{onepose} and the ABO dataset~\cite{abo_dataset}, which include a variety of objects along with their poses. 
Extensive ablation studies highlight the importance of mixing several data sources, and enable to validate our design choices for this relatively novel type of architecture in the field.
We conduct experimental evaluations on the Linemod and OnePose benchmarks. 
Our method outperforms all existing one-shot pose estimation methods %
on the LINEMOD benchmark~\cite{linemod} and performs well in the OnePose benchmark~\cite{onepose}. 
Finally, to demonstrate the robustness of our method in real-world scenarios, we present evaluation results in which a limited number of reference images are provided, outperforming all other methods.

In summary, we make several contributions.
First, we propose a novel transformer-based architecture for object pose estimation that can handle unseen objects at test time without resorting to 3D models. %
Second, we demonstrate the importance of generic 3D-vision pretraining for better generalization in the context of object pose estimation. %
Third, we conduct extensive evaluations and ablations, and show that our method outperforms existing other one-shot methods on several benchmarks.
In particular, our method does not significantly compromise performance in situations with limited information, such as a restricted number of reference images.

\section{Related work}
\label{Related_works}

\PAR{Model-specific approaches}

are only able to estimate the pose of objects for which the method has been specifically trained.
Some of these methods directly regress 6D pose from RGB images~\cite{posecnn, cdpn, pose_guided_auto_enc, GDRnet, lienet}, while others output 2D pixel to 3D point correspondences from which 6D pose can be solved using PnP~\cite{pix2pose, epropnp, pvnet, dpod, bb8, epos}.
In this latter case, most methods leverage accurate CAD models for each object as ground-truth for the 2D-3D mapping~\cite{pix2pose}, and refine pose estimations iteratively~\cite{ssd6d,repose}. 
Although high pose accuracy can be achieved this way, the requirement for exact CAD models hinders scalability and practical use in many application scenarios. 
To eliminate the need for 3D models, recent works~\cite{latentfusion, inerf} use neural rendering models~\cite{nerf} for pose estimation. 
Regardless, model-specific methods remains not scalable, as they need to be retrained for each new object.

\PAR{Category-level methods} learn the shared shape prior within a category and thus eliminate the need for instance-level CAD models at test time~\cite{nocs, shape_category, category_scale_shape, category_recurrent, category_synthesis, fs-net, category_size_space, spga, category_semi_sup_fs, pavllo2023shape}. %
Most of these approaches try to infer correspondences from pixels to 3D points in a Normalized Object Coordinate Space (NOCS). 
Nevertheless, category-level methods still face limitations.
Namely, they can handle only a restricted number of categories and cannot handle objects from unknown categories.

\PAR{Model-agnostic methods} focus on estimating the poses of objects unseen during training, regardless of their category~\cite{bundletrack, fs6d, ove6d, unseen, osop, gen6d, onepose, onepose_plusplus}.
These methods assume that some additional input about the object at hand is provided at test time in order to define a reference pose (otherwise, the pose estimation problem would be ill-defined).
BundleTrack~\cite{bundletrack} and Fs6D~\cite{fs6d}, for instance, requires RGB-D input sequences at inference time. 
More recently, several methods have been proposed for pose estimation of previously unseen objects, given their 3D mesh models. 
For instance, OVE6D~\cite{ove6d} utilizes a codebook to encode the 3D mesh model. 
OSOP~\cite{osop} employs 2D-2D matching and PnP solving techniques based on the 3D mesh model of the object. 
However, these methods require dense depth information, video sequences or 3D meshes that can be challenging to obtain without sufficient time or specific devices. 
This can restrict the use of the models in practical settings.

\PAR{One-shot image-only pose estimation  methods} are a subset of model-agnostic methods that only require minimal input at test time, \ie a set of reference images with annotated poses~\cite{gen6d, onepose, onepose_plusplus}.
Gen6D~\cite{gen6d} uses detection and retrieval to initialize the pose of a query image and then refines it by regressing the pose residual. 
However, it requires an accurate pose initialization and struggles with occlusion scenarios. %
OnePose~\cite{onepose} and OnePose++~\cite{onepose_plusplus} beforehand reconstruct the object 3D point-cloud from the set of reference images using COLMAP~\cite{colmap}, from which 2D-3D correspondences are obtained. 
Although not requiring an explicit 3D mesh models, these two method still need to reconstruct a 3D point-cloud under the hood, which is complex, prone to failure, and not real-time.
In comparison, our method do not need 3D mesh model nor reconstructed point-cloud to infer the object pose.

\section{Method}
\label{Method}

In this section, we first describe the architecture of the proposed model-agnostic approach, then we describe its associated training loss.
Afterwards, we present training details and the 6D pose inference procedure.

\subsection{Model architecture}
\label{sub:model}

Our model takes as input a query image $\I_q$ of the target object $\mathcal{O}$ for which we wish to estimate the pose, and a set of $K$ reference images $\{\I_1,\I_2,\ldots,\I_K\}$ showing the same object under various viewpoints, for which the object pose is known. 
We denote by $\P_i = \{(\R_i, \t_i)\}$ the pose of the object relatively to the camera in the reference image $\I_i$.
Here we assume prior knowledge of the object instance in the query image, which is typically provided by an object detector or a retrieval system applied beforehand. 
For the sake of simplicity and without loss of generality, we also assume that all images (query and reference images) are approximately cropped to the object bounding box.

\PAR{Overview of the architecture.}
Figure~\ref{fig:overview} shows an overview of the model architecture. 
First, the query and reference images are encoded into a set of token features with a Vision Transformer (ViT) encoder~\cite{vit}. 
For each reference image, the object pose is then encoded and injected into the image features using cross-attention. %
This latter module, to which we refer as \emph{pose encoder}, outputs visual features \emph{augmented} with 6D pose information.
A transformer decoder then jointly process the information from the query features with the augmented reference image features.
Finally, a prediction head outputs dense 3D coordinates for each pixel of the query image, from which we can recover the 6D pose in the query image.
We now describe each module in details. %

\PAR{Image encoder.}
We use a vision transformer~\cite{vit} to encode all query and database images.
In more details, each image is divided into non-overlapping patches, and a linear projection encodes them into patch features. 
A series of transformer blocks is then applied on these features: each block consists of multi-head self-attention and a MLP. 
In practice, we use a ViT-Base model, \ie $16{\times}16$ patches with $768$-dimensional features, $12$ heads and $12$ blocks.
Following~\cite{deepmatcher,crocostereo}, we use RoPE~\cite{rope} relative position embeddings.
As a result of the ViT encoding, we obtain sets of token features denoted $\F_q$ for the query and $\F_i$ for the reference image $\I_i$ respectively:
\begin{equation}
\begin{cases}
\F_q = \text{ImageEncoder} \left( \I_q \right), \\
\F_i = \text{ImageEncoder} \left( \I_i \right), i=1 \ldots K.
\end{cases}
\end{equation}

\begin{figure}
    \centering 
    \hspace{-3mm}
    \centering
    \includegraphics[width=.15\textwidth]{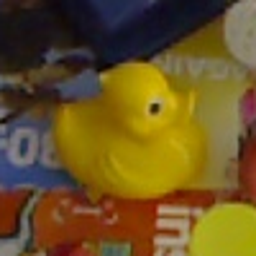} 
    \includegraphics[width=.15\textwidth]{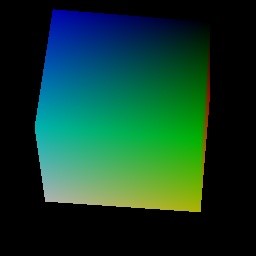}
    \includegraphics[width=.15\textwidth]{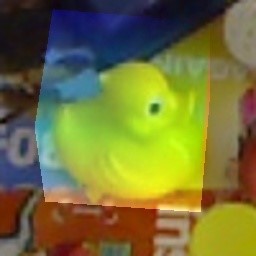}
    \caption{
    \textbf{Reference view and its associated proxy shape.} 
    Illustration of a cuboid proxy shape used to jointly represent the object pose and dimension in a friendly data-format for transformers.
    The proxy shape is rendered into a dense 3D coordinate map \wrt the object coordinate system, represented here as a 3-channel image.
    }
    \label{fig:ref_proxy}
    \vspace{-5mm}

\end{figure}

\begin{figure}[ht!]
    \centering 
    \hspace{-3mm}
    \centering
    \includegraphics[trim=0 295 253 0,clip,width=\linewidth]{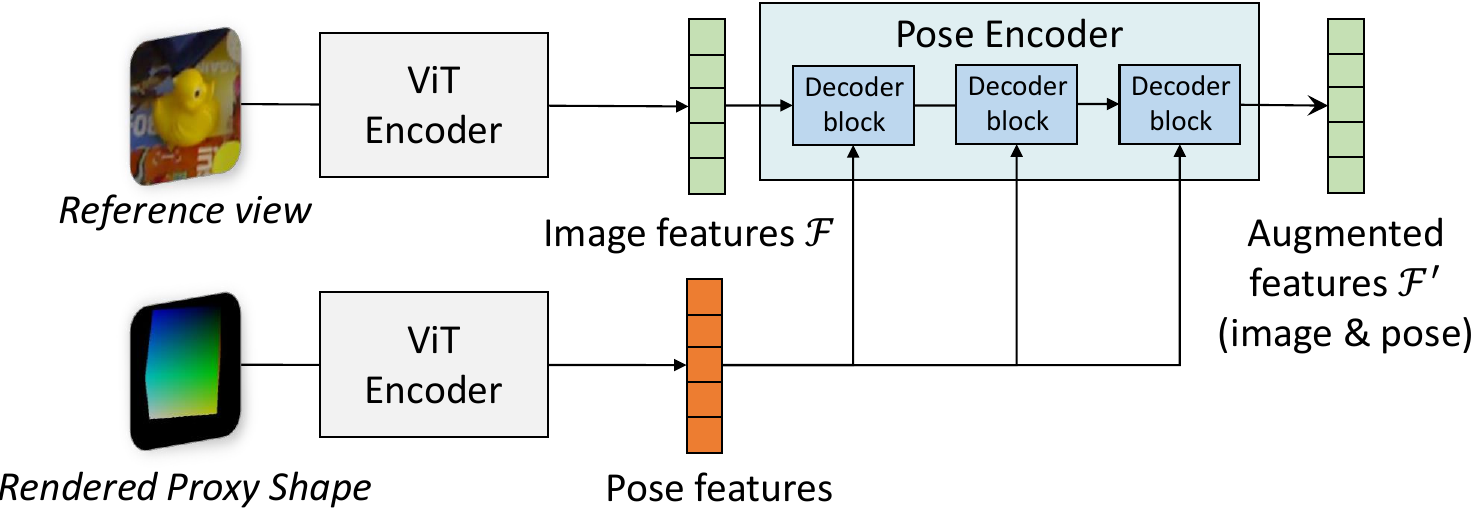}
    
    \caption{
    \label{fig:posenc}
    \textbf{Architecture of the Pose Encoder.} 
    The pose encoder combines the reference image features $\F$ with the annotated object pose, in the form of a rendered 3D proxy shape, yielding the pose-augmented features $\Faug$. 
    }

\end{figure}

\PAR{Pose encoder.}
There are multiple ways of inputting a 6D pose $\P_i$ to a deep network, see~\cite{roma}.
Since we aim to combine the 6D object pose with its visual representation $\F_i$, we opt for an image-aligned pose representation which blends seamlessly with the visual representation $\F_i$.
Specifically, as shown in Figure~\ref{fig:ref_proxy}, %
we transform the pose into an image by rendering 3D coordinates of a proxy shape (\eg a cuboid or an ellipsoid), scaled according to the object dimension, and positioned according to the 6D object pose.
As illustrated in Figure~\ref{fig:posenc}, this 3-channel image is fed to another ViT, and then mixed with the visual features $\F_i$ through the cross-attention layers of transformer decoder, yielding the pose augmented features $\Faug_i$:
\begin{equation}
    \Faug_i = \text{PoseEncoder} \left( \F_i, \text{ViT}(\text{Render}(\P_i)) \right) .
\end{equation}

\begin{figure}
    \centering
    \includegraphics[width=0.49\linewidth]{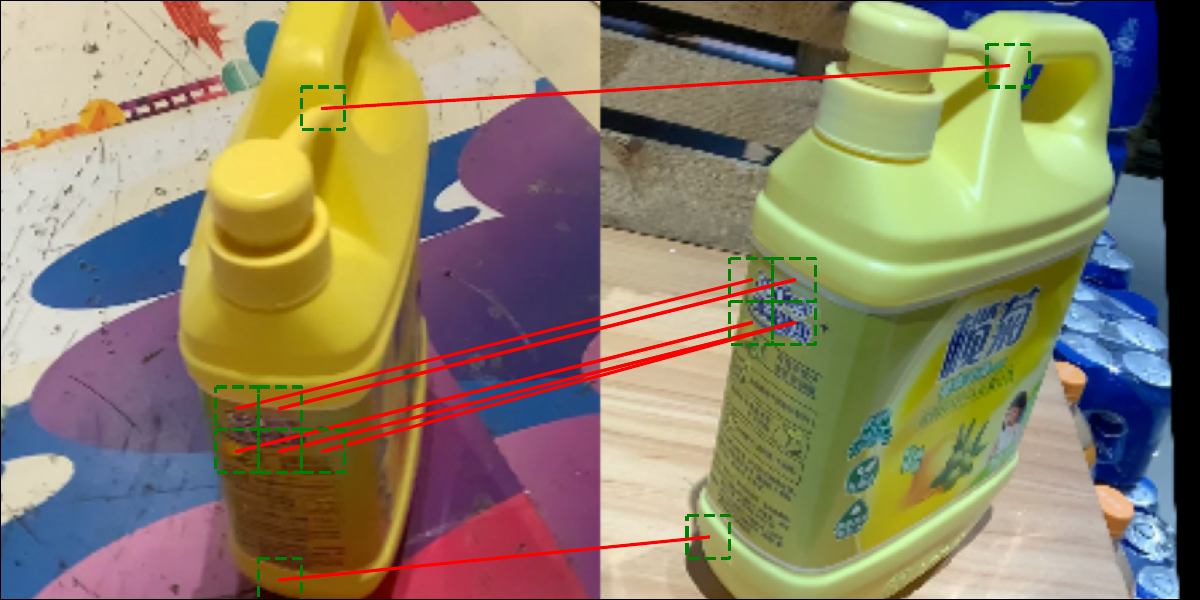}
    \includegraphics[width=0.49\linewidth]{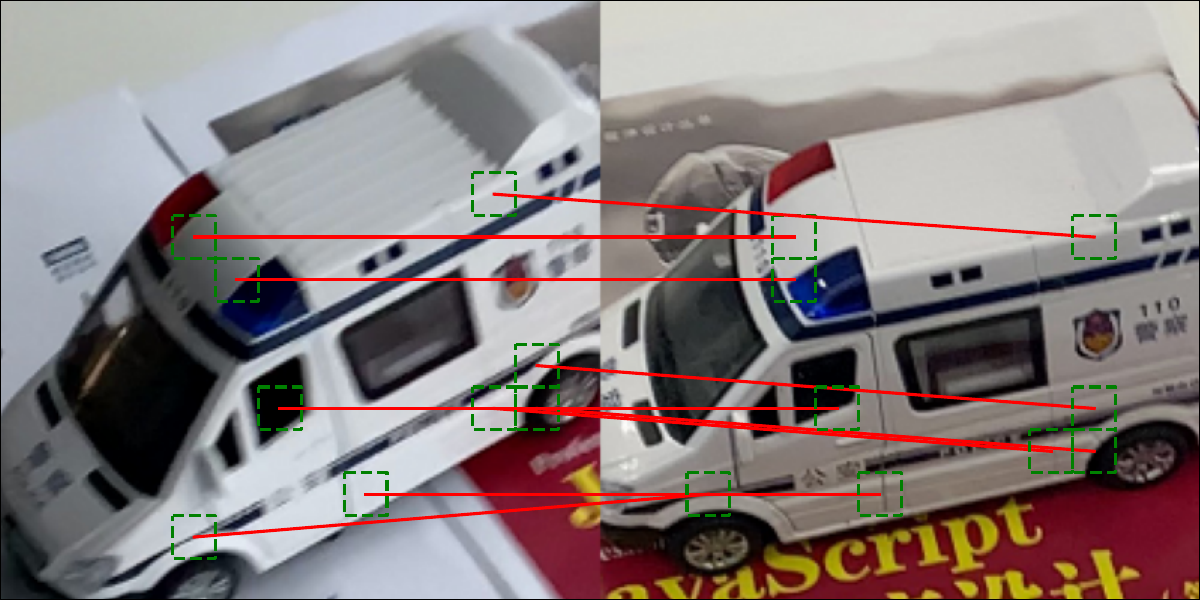}
    \caption{Visualization of the cross-attention in the decoder.
        Here we plot the top-10 attentions as correspondences between 2 tokens (\ie 16x16 patches) in the query and reference image, respectively.}
    \label{fig:cross_attention}
    \vspace{-5mm}
\end{figure}

\PAR{Decoder.}
The next step is to extract relevant information from the reference images with respect to the query image (not all reference images are necessarily helpful). 
To that aim, we again leverage a transformer decoder that compares the query features $\F_q$ to all concatenated tokens ${\Faug_i}$ from the augmented reference images via cross-attention.

\PAR{Prediction head.}
After obtaining the token features from the last transformer decoder block, 
we project them using a linear head and reshape the result as a 4-channel image with the same resolution as the query image.
For each pixel, we thus predict the 3D coordinates of the associated point on the proxy shape, and an additional 4th channel yields the confidence $\tau$ (see below).
Note that we predict the 3D coordinates on the surface of the proxy shape, not those on the surface of the target object.
Finally, a robust PnP estimator extracts the most likely pose from this predicted 2D-3D mapping.

\subsection{Training losses}
\label{sub:loss}

\PAR{3D regression loss.} 
A straightforward way to train the network is, for each pixel $i$, to regress the ground-truth 3D coordinates of the proxy shape at this pixel.
We use an Euclidean loss for pixels where such ground-truth is available:
\begin{equation}
    \Lregr^{(i)} = \left\Vert \yh_i - \y_i^{gt} \right\Vert,
    \label{eq:loss_reg}
\end{equation}
where $\yh_i \in \mathbb{R}^3$ is the network output for the $i^{th}$ pixel of the query image, and $\y_i^{gt}$ is the corresponding ground-truth 3D point.

\PAR{Pixelwise confidence.}
Since it is unlikely that all pixels can get correctly mapped during inference (\eg background pixels), it is important to assert the likelihood of correctness of the predicted 2D-3D mapping for each pixel.
Following~\cite{kendall2018multi}, we therefore jointly predict a per-pixel indicator $\tau_i > 0$ to modulate the pixelwise loss $\Lregr$ to form the final loss as follows:
\begin{equation}
    \Lfinal^{(i)} = \tau_i \Lregr^{(i)} - \log \tau_i.    
    \label{eq:loss_conf}
\end{equation}
Note that $\tau$ can be interpreted as the confidence of the prediction, as if $\tau_i$ is low for pixel $i$, the corresponding error $ \mathcal{L}^{(i)}$ at this location will be down-weighted, and \textit{vice versa}. 
For pixels outside the proxy shape, we set $\Lregr^{(i)}=E$, where $E$ is a constant representing a large regression error. %
The second term of the loss acts as a regularizer, so as to prevent the model from getting under-confident everywhere.

\subsection{Training details}
\label{sub:details}

\PAR{Training data.}
To ensure the generalization capability of our model, we train it on a diverse set of datasets covering a large panel of diversity. 
Specifically, we choose the large-scale ABO dataset~\cite{abo_dataset}, which comprises 580K images from 8,209 sequences, featuring 576 object categories (mostly furniture) from amazon.com. 
We also use for training some datasets of the BOP challenge~\cite{bop_dataset}, namely T-LESS, HB, HOPE, YCB-V, RU-APC, TUD-L, TYO-L and ICMI.
We exclude symmetrical objects from the training set, as well as 3 objects from the HB dataset which exist in the LINEMOD dataset, in order to evaluate our generalization capabilities on this benchmark. 
In total, we consider 150K synthetic physically-based-rendered images and 53K real images, featuring 153 objects, from the BOP challenge for training. 
Additionally, we incorporate the OnePose dataset~\cite{onepose}, which includes over 450 video sequences of 150 objects captured under various background environments.

To get the dimensions of the proxy shape, we either use the convex hull of the 3D mesh (if available), otherwise we use the 3D bounding box.
Note that the exact dimension and orientation of the proxy shape have little impact on the final performance, as demonstrated in Supplementary material.

\PAR{Memory optimization.}
During training, we feed the network with batches of $16\times48=768$ images, each batch being composed of 16 objects for which 16 query and 32 reference images are provided (48 images in total).
Since queries of the same object attend to the same set of reference images,
we precompute features $\{\Faug\}$ for these reference images and share them across all queries.
Furthermore, by a careful reshaping of the tensors in-place in the query decoder, we can resort to vanilla attention mechanisms without any copy in memory (see Supplementary material for details).
In addition to considerably reducing the memory requirements, this optimization significantly speeds up training. 
 
\PAR{Network architecture and training hyper-parameters.}
We use a ViT-Base/16~\cite{vit} for the image encoder.
The decoder is identical, except it has additional cross-attention modules.
For the pose encoder, we use a single-layer ViT to encode the proxy shape rendering, and $4$ transformer decoder blocks to inject the pose information into the visual representation.
We use relative positional encoding (RoPE~\cite{rope}) for all multi-head attention modules.
We train our network for 40,000 steps with AdamW with $\beta=(0.9,0.95)$ and a cosine-decaying learning rate going from $10^{-4}$ to $10^{-6}$.
We initialize the network weights using CroCo v2~\cite{crocostereo}, a recently proposed pretraining method tailored to 3D vision and geometry learning.
We evaluate the impact of geometric pretraining in the next section.

\PAR{Data augmentation.}
We recale and crop all images to a resolution of $224 \times 224$ around the object location. %
We apply standard augmentation techniques for cropping, such as random shifting, scaling and rotation to increase the diversity of our training data. 
We also apply augmentation to the input of our \emph{pose encoder} to improve generalization. 
We specifically apply random geometric 3D transformations to the proxy shape pose and coordinates, including 3D rotation, translation and scaling.
When choosing the set of 32 reference images for each object, we select 8 reference images at random across the entire pool of reference images for this object, and the remaining 24 views are selected using a greedy algorithm,\ie farthest sampling that minimizes blind spots.

\subsection{Inference procedure}
\label{sub:inference}

\PAR{3D proxy shape.}
Our pose encoder receives multiple reference images of the target object and their corresponding 6D poses. 
Given a proxy shape template (\eg a cuboid or an ellipsoid), we first align the 3D proxy shape centroid with the object center (according to the ground-truth pose). 
We then scale the proxy shape according to the target object dimensions. 
The generated 3D proxy shape is then transformed according to the object pose and rendered to the camera, yielding a 3D point map, see Figure~\ref{fig:ref_proxy}.

\PAR{Predicting object poses.}
To solve the object pose in a given query image, we sample $K$ reference views among all the available reference views for this object.
We use a greedy algorithm~\cite{farthest_point_sampling} to maximize the diversity of viewpoints in the selected pool of views.
From this input, our model predicts a dense 2D-3D mapping and an associated confidence map, as can be seen in Figure~\ref{fig:overview}. 
We then filter out regions for which the confidence is below a threshold $\tau$.
Finally we use an off-the-shelf PnP solver to obtain the predicted object pose. 
Specifically, we rely on SQ-PnP~\cite{sqpnp} with 1024 randomly sampled 2D-3D correspondences, according to the confidence of the remaining points, maximum 1000 iterations and a reprojection error threshold of 5 pixels. %

\section{Experiments}
\label{experiments}

\subsection{Dataset and metrics}

\PAR{Test benchmarks.}
We use the test splits of the training datasets explained earlier.
In more details, we evaluate on the LINEMOD~\cite{linemod} dataset, a subset of the BOP benchmark~\cite{abo_dataset}, a widely-used dataset for object pose estimation comprising 13 models and 13K real images.
For the evaluation, we use the standard train-test split proposed in~\cite{cdpn} and follow the protocol defined in OnePose++~\cite{onepose_plusplus}, using their open-source code and detections from the off-the-shelf object detector YOLOv5~\cite{yolov5}. 
In more details, we use approximately 180 real training images as references, discarding the 3D CAD model and only using the pose, while all remaining test images are used for evaluation. 
For the Onepose~\cite{onepose} and ABO~\cite{abo_dataset} datasets, we use the official test splits as well.
We also use OnePose-LowTexture dataset~\cite{onepose_plusplus}, where there are 40 household low-textured objects for evaluation.

\PAR{Metrics.}
We use the \emph{cm-degree} metric to evaluate the accuracy of our predicted poses on both datasets. 
The rotation and translation errors are calculated separately, and a predicted pose is considered correct if both its rotation error and translation error are below a certain threshold. 
For the LINEMOD dataset, CAD models are available to evaluate the accuracy, and therefore, we employ two additional metrics: the \emph{2D projection metric} and the \emph{ADD metric}. 
We set the threshold for the \emph{2D projection metric} to 5 pixels. 
To compute the \emph{ADD metric}, we transform the 3D model's vertices using both the ground truth and predicted poses, and calculate the average distance between the two sets of transformed points. 
We consider a pose accurate if the average pointwise distance is smaller than 10\% of the object's diameter. 
For symmetric objects, we consider the average point-to-set distance (\emph{ADD-S}) instead~\cite{posecnn}.

\begin{table*}[ht!]
    \centering
    \setlength\tabcolsep{3pt} %
    \resizebox{0.95\textwidth}{!}{
    \begin{tabular}{c| c c c c c c c c c c c c c |c} %
    
    \toprule %
        \multirow{2}*{Name} & \multicolumn{13}{c|}{Object Name} & \multirow{2}*{Avg.} \\
        & ape & benchwise & cam & can & cat & driller & duck & eggbox${}^{*}$ & glue${}^{*}$ & holepuncher & iron & lamp & phone & \\ 
        \hline
        \multicolumn{1}{c|}{} & \multicolumn{13}{c|}{\textit{ADD(S)-0.1d}} &  \\ \hline

        Gen6D & - & 62.1 & 45.6 & - & 40.9 & 48.8 & 16.2 & - & - & - & - & - & - & -\\  
        OnePose & 11.8 & 92.6 & 88.1 & 77.2 & 47.9 & 74.5 & 34.2 & 71.3 & 37.5 & 54.9 & 89.2 & 87.6 & 60.6 & 63.6\\
        OnePose++ & 31.2 & 97.3 & 88.0 & 89.8 & 70.4 & 92.5 & 42.3 & 99.7 & 48.0 & 69.7 & 97.4 & 97.8 & 76.0 & 
        76.9\\
        \textbf{Ours ($K=16$)} &  39.4&      64.6&      73.1&      76.3&      63.0&      83.5&      43.4&      99.2&      61.3&      83.7&      72.1&      84.1&      45.1&      68.4\\
        \textbf{Ours ($K=64$)} & 47.2&      73.5&      87.5&      85.4&      80.2&      92.4&      60.8&      99.6&      69.7&      93.5&      82.4&      95.8&      51.6&     \bf 78.4\\
        
        \midrule %

        \multicolumn{1}{c|}{} & \multicolumn{13}{c|}{\textit{Proj2D}} & \\ \hline
       
        OnePose & 35.2 & 94.4 & 96.8 & 87.4 & 77.2 & 76.0 & 73.0 & 89.9 & 55.1 & 79.1 & 92.4 & 88.9 & 69.4 & 78.1\\ 
        OnePose++ & 97.3 & 99.6 & 99.6 & 99.2 & 98.7 & 93.1 & 97.7 & 98.7 & 51.8 & 98.6 & 98.9 & 98.8 & 94.5 & 94.3 \\
        \textbf{Ours ($K=16$)} & 96.6&      82.9&      95.1&      92.7&      95.4&      89.9&      89.4&      98.6&      94.0&      98.5&      79.1&      85.2&      76.0&      90.3\\
        \textbf{Ours ($K=64$)} & 97.1&      94.1&      98.4&      98.2&      98.4&      95.7&      96.3&      99.0&      94.8&      99.3&      94.6&      94.2&      88.9&     \bf 96.1 \\
        \bottomrule %
        \end{tabular}
    }
    \caption{\textbf{Results on LINEMOD} 
    and comparison with other \textit{one-shot} baselines. Symmetric objects are indicated by ${}^*$. %
    }
    \label{tab:exp_linemod}
    \vspace{-3mm}
\end{table*}

\subsection{Ablative study}

We first conduct several ablative studies to measure the impact of critical components in our method, such as the choice of the proxy shape, training data and pretraining, or the number of reference images.
For these experiments, we report numbers on subsets of the three benchmarks mentioned above.
We uniformly sample 5000 queries from ABO dataset and report each time a few adequate metrics. %
Unless specified otherwise, we use the same training sets, hyper-parameters and network architecture specified previously.

\PAR{Impact of different proxy shapes.}
We first experiment with two simple proxy shapes: a cuboid or an ellipsoid.
As shown in Table~\ref{tab:ab_proxy}, using the cuboid proxy shape yields superior performance consistently on all datasets.
To get more insights, we also try to predict 3D coordinates aligned with the object's surface, \ie we try to predict the CAD model given cuboid proxy shapes as input reference poses. 
In this case, we exclude the OnePose dataset from the training set, since no CAD model is available.
Interestingly, this cuboid-to-CAD setting performs much worse than cuboid-to-cuboid, meaning that it is easier for the network to regress 3D coordinates of an invisible cuboid (not necessarily aligned with the object surface) than actually reconstruct the object's unknown 3D shape.
In other words, the model \emph{does not need} to know nor infer the 3D object shape to estimate its pose.
We use the cuboid-to-cuboid setting in all subsequent experiments.

\begin{table}[ht]
    \centering
    \resizebox{\linewidth}{!}{
    {\small
    \setlength\tabcolsep{2pt} %
    \begin{tabular}{c|c|c|c} 
    \toprule
    \multirow{2}{*}{Proxy shape (input$\rightarrow$output)} & \multicolumn{1}{c|}{LINEMOD} & \multicolumn{1}{c|}{OnePose} & \multicolumn{1}{c}{ABO} \\ 
      & \small ADD(s)$\uparrow$ & \small 5cm-5deg~$\uparrow$ & \small 5cm-5deg~$\uparrow$ \\ 
    \midrule
    ellipsoid $\rightarrow$ ellipsoid  & 58.0 & 79.9& 70.8 \\
    cuboid $\rightarrow$ cuboid  & \bf 60.9 & \bf 88.3 & \bf 74.4 \\
    \midrule
    cuboid $\rightarrow$ CAD model   &  42.3& 40.4 & 63.7 \\
    \bottomrule
    \end{tabular}
    }}
    \caption{\textbf{Impact of the 3D proxy shape.}
    }
    \label{tab:ab_proxy}
\end{table}

\PAR{Training data ablation.}
We then conduct an ablation to measure the importance of diversity in the training data.
To that aim, we discard parts of the training set, still ensuring that all models trains for the same number of steps in each setting for the sake of fair comparison.
Table~\ref{tab:ab_dataset} shows that having more diversity in the training set is critical to improve performance on all test sets.
This result suggests that, despite the great diversity between datasets (for instance, ABO contains mostly furnitures), knowledge can effectively be shared and transferred between datasets. %

\begin{table}[ht]
    \centering
    {\small
    \setlength\tabcolsep{2pt} %
    \begin{tabular}{c|c|c|c} 
    \toprule
    \multirow{2}{*}{Training Dataset} & \multicolumn{1}{c|}{LINEMOD} & \multicolumn{1}{c|}{OnePose} & \multicolumn{1}{c}{ABO} \\ 
      & \small  ADD(s)$\uparrow$ & \small 5cm-5deg~$\uparrow$ & \small 5cm-5deg~$\uparrow$ \\ 
    \midrule
    BOP & 44.2 & 72.0 & 3.44  \\
    BOP + OnePose & 49.5 & 83.2 & 5.62 \\
    BOP + OnePose + ABO  & \bf 60.9 & \bf 88.3 & \bf 74.4 \\
    \bottomrule
    \end{tabular}
    }
    \caption{\textbf{Ablation on training datasets.}
    }
    \label{tab:ab_dataset}
\end{table}

\PAR{Impact of the number of reference images.}
We measure the effect of varying at test time the number of reference views $K$, which go through the \emph{Pose encoder} as input. As shown in Table~\ref{tab:ab_ref_views}, increasing the number of reference views lead to higher performance, because of the increased number of views potentially closer to the query viewpoint. 
This is important for practicality, because it shows that a model trained with a certain number of reference views at train time can handle a different number of reference views at test time.

\begin{table}[ht!]
    \centering
    {\small
    \setlength\tabcolsep{2pt} %
    \begin{tabular}{c|c|c|c|c} 
    \toprule
    \multirow{2}*{\textbf{Train}} & \multirow{2}{*}{\textbf{Test}} &  \multicolumn{1}{c|}{LINEMOD} & \multicolumn{1}{c|}{OnePose} & \multicolumn{1}{c}{ABO} \\ 
      & & \small ADD(s)$\uparrow$ & \small 5cm-5deg~$\uparrow$ & \small 5cm-5deg~$\uparrow$\\ 
    \midrule
    \multirow{3}*{{$K=16$}} & $K = 16$ & 60.9 & 88.3 & 74.4 \\
    & $K = 32$ & 63.8 & 88.8 & 75.0 \\
    & $K = 64$ & 65.3 & \bf 89.0 & 75.4 \\
    \midrule
    \multirow{3}*{{$K=32$}} & $K = 16$ & 68.4 & 87.8 & 74.8 \\
    & $K = 32$ & 75.5 & 88.4 & 76.9 \\
    & $K = 64$ & \bf 78.4 & 88.6 & \bf 77.0 \\
    
    \bottomrule
    \end{tabular}
    }
    
    \caption{\textbf{Impact of the number of reference images at train and test time.}
    }
    \label{tab:ab_ref_views}
    \vspace{-5mm}

\end{table}

\PAR{Impact of pretraining.}
We finally assess the benefit of preemptively pretraining the network with a self-supervised objective.
We specifically investigate whether pretraining is beneficial, and in particular, whether it should be \emph{geometrically-oriented} or not.
We thus compare CroCo pretraining~\cite{croco} with MAE pretraining~\cite{MAE}.
The latter yields state-of-the-art results in many vision tasks, and is in addition compatible with our ViT-based architecture.
Contrary to CroCo, however, MAE has no explicit relation to 3D geometry. 
We present results in Table~\ref{tab:ab_pretrained}.
We first note a considerable drop in performance when the network is trained from scratch (\ie no pretraining). 
We then observe that, while MAE pretraining does improve a lot over no pretraining at all, it is still largely behind the performance attained by CroCo pretraining.
Note that there is no unfair advantage in using CroCo, since CroCo is \emph{not} trained on any object-centric data. 
Rather, CroCo pretraining data includes scene-level and landmark-level indoor and outdoor scenes, such as Habitat, MegaDepth, etc. (see~\cite{crocostereo} for the complete dataset list).
Note that we systematically measure generalization performance (\ie testing on unseen objects), hence clearly demonstrating how geometry-oriented pretraining is crucial for generalization.

\begin{table}[ht!]
    \centering
    \resizebox{\linewidth}{!}{
    {\small
    \setlength\tabcolsep{2pt} %
    \begin{tabular}{c|cc|cc} 
    \toprule
    \multirow{2}{*}{ pretraining scheme} & \multicolumn{2}{c|}{LINEMOD} & \multicolumn{2}{c}{OnePose} \\ 
      & \small ADD(S)-0.1d~$\uparrow$ & \small Proj2D~$\uparrow$  & \small 3cm-3deg~$\uparrow$ & \small 5cm-5deg~$\uparrow$\\ 
    \midrule
    None  & 16.6 & 27.3 & 27.5 & 54.8 \\
    MAE~\cite{MAE} & 39.4 & 56.7 & 54.0 & 72.3 \\
    Croco~\cite{crocostereo} & \bf 68.4 & \bf 90.3 & \bf 76.3 & \bf 87.8 \\

    \bottomrule
    \end{tabular}
    }}
    \caption{\textbf{Impact of the pre-training strategy.}
    }
    \label{tab:ab_pretrained}
\end{table}

\begin{table*}[ht!]
    \centering
    \resizebox{0.59\textwidth}{!}{
    \setlength\tabcolsep{2pt} %
    \begin{tabular}{c|c|cc|ccc|ccc} 
    \toprule
    \multirow{2}{*}{} & \multirow{2}{*}{$K$} &  \multicolumn{2}{c|}{LINEMOD} &  \multicolumn{3}{c|}{OnePose dataset} & \multicolumn{3}{c}{OnePose-LowTexture}  \\ 
    & & \small ADD(s)$\uparrow$ & \small Proj2D~$\uparrow$ & {\small 1cm-1deg}  & \small 3cm-3deg  & \small 5cm-5deg & \small 1cm-1deg & \small 3cm-3deg & \small 5cm-5deg  \\ 
    \midrule

    OnePose++ & \multirow{2}{*}{$8$} & 10.3 & 10.4 & \bf 36.1 & 62.4 & 67.9 & 4.2 & 13.9 & 18.5 \\
    Ours & & \bf 55.5 & \bf 75.9 & 25.0 & \bf 72.6 & \bf 85.7 & \bf 9.7 & \bf 44.8 & \bf 65.2 \\  
    
    \midrule
    OnePose++ & \multirow{2}{*}{$16$} & 35.2 & 57.9 & \bf 46.6 & 76.1 & 82.8 & 12.1 & 39.2 & 51.6 \\
    Ours & & \bf 68.4 & \bf 90.3 & 28.5 & \bf 76.3 & \bf 87.8 & \bf  12.4 & \bf 51.3 & \bf 71.9\\
    \midrule
    OnePose++ & \multirow{2}{*}{$32$}  & 56.7 & 82.1 & \bf 49.7 & \bf 78.6 & 85.4 & \bf 16.8 & 52.9 & 67.0  \\
    Ours & & \bf 75.5 & \bf 94.7 & 29.6 & 77.6 & \bf 88.4 & 14.1 & \bf 53.6 & \bf 73.4 \\    
    \midrule
    OnePose++ & \multirow{2}{*}{$64$}  & 56.8 & 90.2 & \bf 50.6 & \bf 80.0 & 86.6 & \bf 16.8 & \bf 56.2 & 71.1  \\
    Ours & & \bf 78.4 & \bf 96.1 & 30.0 & 78.0 & \bf 88.6 & 14.1 & 54.3 & \bf 74.2 \\
    \midrule
    OnePose++ & All & 76.9 & 94.3 & 51.1 & 80.8 & 87.7& 16.8 & 57.7 & 72.1 \\
    \bottomrule
    \end{tabular}
    }
    \caption{\textbf{Comparison of our model and OnePose++ with restricted numbers of reference images $K$.}
    }
    \label{tab:ab_ref_views_lm_onepose}
    \vspace{-3mm}

\end{table*}

\PAR{Visualization.}
To understand how the network works internally, we visualize interactions happening in the cross-attention of the decoder in Figure~\ref{fig:cross_attention}.
Undeniably, the model does perform matching under the hood to solve the task, as we see that all interactions consist of token-level correspondences between their corresponding patches. 
This is interesting, because the network is never explicitly trained for establishing correspondences.
This also explains why the CroCo pretraining is so important, as this latter essentially consists in learning to establish correspondences between different viewpoints, see~\cite{croco}.

\subsection{Comparison with the state of the art}

\PAR{LINEMOD.}
We compare against Gen6D~\cite{gen6d}, OnePose~\cite{onepose} and OnePose++~\cite{onepose_plusplus}, which are one-shot methods similar to our approach on the \textit{ADD(S)-0.1d} and \textit{Proj2D} metrics. %
As shown in Table~\ref{tab:exp_linemod}, our approach outperforms these one-shot methods. 
Compared to the other one-shot methods, it is noteworthy that our method does not require any knowledge of the 3D object shape as input, in contrast to OnePose and OnePose++ which reconstruct 3D SfM model in advance.
Our method gives 1.5\% and 1.8\% improvements on the ADD-S and Proj2D metrics, respectively, compared to the best competitor.

\PAR{OnePose and OnePose-LowTexture.}
We again compare our approach with OnePose and OnePose++~\cite{onepose,onepose_plusplus}, as well as some SfM baselines, on the challenging OnePose test set, which has the particularity of not providing CAD models.
Results are provided in Table~\ref{tab:exp_onepose} in terms of the standard \emph{cm-degree} accuracy for different thresholds.
Note that ``HLoc (LoFTR${}^{*}$)'' uses LoFTR coarse matches for SfM and uses full LoFTR to match the query image and its retrieved images for pose estimation. 
Our method lags behind OnePose++ at the tightest 1cm/1deg threshold. 
In contrast to methods based on establishing pixel correspondences, such as OnePose++, which can be pixel-precise as a matter of fact, and therefore provide high-precision pose estimates, our method predicts the coordinates of an `invisible' proxy shape. 
This is definitely harder, and as a result, the resulting pose estimate is noisier. 
However, as the accuracy threshold of the performance metric increases (5cm/5deg), our method outperforms correspondence-based methods, demonstrating better robustness overall to challenging conditions.

\begin{table}[ht!]
    \centering
    \resizebox{\linewidth}{!}{
    \setlength\tabcolsep{2pt} %
    \begin{tabular}{c|c|ccc|ccc} 
    \toprule
    \multirow{2}{*}{} &  & \multicolumn{3}{c|}{OnePose dataset} & \multicolumn{3}{c}{OnePose-LowTexture}  \\ 
     & SfM & {\small 1cm-1deg}  & \small 3cm-3deg  & \small 5cm-5deg & \small 1cm-1deg & \small 3cm-3deg & \small 5cm-5deg  \\ 
    \midrule
    HLoc~\textit{(SPP + SPG)} & yes & 51.1 & 75.9  & 82.0 & 13.8 & 36.1 & 42.2  \\
    HLoc~\textit{(LoFTR${}^{*}$)} & yes & 39.2 & 72.3 & 80.4 & 13.2 & 41.3 & 52.3  \\
    OnePose & yes & 49.7 & 77.5 & 84.1 & 12.4 & 35.7 & 45.4 \\
    OnePose++ & yes & \bf 51.1 & \bf 80.8 & 87.7 & \bf 16.8 & \bf 57.7 & 72.1 \\
    \midrule
    \textbf{Ours ($K=16$)} & no & 28.5 & 76.3 &  87.8 & 12.4 & 51.3 & 71.9 \\
    \textbf{Ours ($K=64$)} & no & 30.0 & 78.0 & \bf 88.6 & 14.1 & 54.3 & \bf 74.2 \\

    \bottomrule
    \end{tabular}
    }
    \caption{\textbf{Comparison with \textit{One-shot} Baselines.}
    Our method is compared with HLoc~\cite{hfnet} combined with different feature matching methods~\cite{superglue, loftr}, OnePose~\cite{onepose} and OnePose++~\cite{onepose_plusplus}.
    We denote as `SfM' methods relying on an explicit 3D reconstruction of the objects.
    }
    \label{tab:exp_onepose}

\end{table}

\PAR{Limited number of reference images.}
We also compare with OnePose++ in scenarios where the number of available reference images is limited.
We experiment with various settings by altering the number of reference images ($K$) and report results in Table~\ref{tab:ab_ref_views_lm_onepose}.
In the case of OnePose++, the `All' configuration entails using 170 and 130 reference images on average on the LINEMOD and OnePose datasets, respectively. %
It is noteworthy that as $K$ decreases to values below 32, the performance of OnePose++ significantly drops on both LINEMOD and OnePose-LowTexture datasets. 
In contrast, our method exhibits a steady performance with only marginal degradation in accuracy. 
This result demonstrates the superior robustness of our approach in situations where the number of available reference images is limited.

We point out that our method is more practical than OnePose++, since it indiscriminately takes videos or small image sets with camera poses as raw inputs. 
In comparison, OnePose++ relies on videos and SfM pre-processing to build 3D object representations (we note they also rely on ground-truth poses from ARKit-scene), which is slow, complex and prone to failure -- all of this strongly impairing scalability.

\subsection{Detailed timings}

We report in Table~\ref{tab:inference_time} inference timings for a single query image, and assuming that reference views have been pre-encoded offline, measured on a single V100 GPU (repeating experiments 10 times and keeping the median timings):

\begin{table}[ht!]
\begin{center}
\begin{tabular}{c|c| c}
    \toprule
    Step & Time ($K=16$) & Time ($K=64$)\\
    \midrule
    {Image encoder} & 3.40 ms & 3.40 ms \\
    {Decoder}& 17.21 ms & 58.79 ms\\
    {Linear head}& 0.05 ms & 0.05 ms \\
    \midrule
    Total & 20.66 ms & 62.24 ms \\
    \bottomrule
    
\end{tabular}
\caption{\textbf{Timing of our method.} Our method is 3\textasciitilde 4 faster than OnePose~\cite{onepose} and OnePose++~\cite{onepose_plusplus} whose 2D-3D matching modules take 66.4 and 88.2ms respectively on a single V100 GPU.
}
\label{tab:inference_time}
\vspace{-5mm}

\end{center}
\end{table}

\section{Conclusion}

We propose a novel approach, called MFOS, for model-free one-shot object pose estimation. 
In contrast to existing one-shot methods, MFOS does not need any 3D model of the target object, such as a mesh or point-cloud, and only requires a set of reference images annotated with the object poses and its approximate size. 
It is able to implicitly extract 3D information from reference images, jointly matching, combining and extrapolating pose information with the query image, using only generic modules from a ViT architecture.
In contrast to all existing methods, our approach is inherently simple, practical and scalable.
In an extensive ablative study, we have determined good practices with this novel type of architecture in the field.
Experiments show that our approach outperforms existing one-shot methods and show significant robustness in scenarios with a limited number of reference images.

\bibliography{aaai24}

\appendix
\pagebreak
{\Huge \textbf{Appendix}}
\\
\\

This appendix provides additional information to the submission \emph{MFOS: Model-Free \& One-Shot object pose estimation}.
We provide qualitative results and additional numerical evaluations for our method.
We also explain implementation details, including general and ethical considerations regarding this work.

\section{Qualitative results}
\label{sec:qualitative_results}

\begin{figure*}
\begin{center}
    \captionsetup{type=figure}
    \resizebox{\linewidth}{!}{
    \begin{SCRfigtabular}
    \SCRfigline{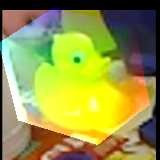} \\
    \SCRfigline{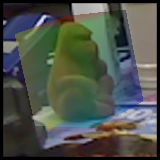} \\
    \SCRfigline{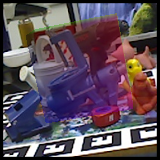} \\
    \SCRfigline{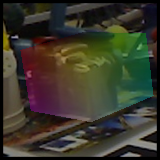} \\
    \SCRfigline{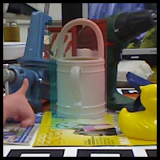} \\
    \SCRfigline{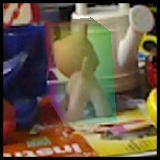} \\
    \SCRfigline{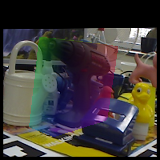} \\
    \SCRfigline{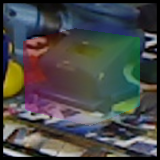} \\
    \SCRfigline{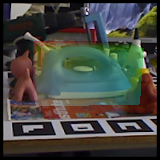} \\
    \end{SCRfigtabular}}
    \\[-0.2cm]
    \captionof{figure}{\textbf{Regression examples} on the LINEMOD dataset~\cite{linemod}.  Best viewed in color.}
    \label{fig:regression_results_linemod2}
\end{center}
\end{figure*}

\begin{figure*}
\begin{center}
    \captionsetup{type=figure}
    \resizebox{\linewidth}{!}{
    \begin{SCRfigtabular}
    \SCRfigline{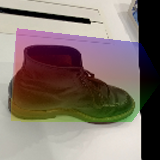} \\
    \SCRfigline{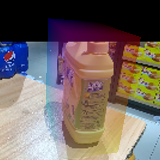} \\
    \SCRfigline{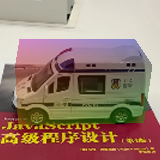} \\
    \SCRfigline{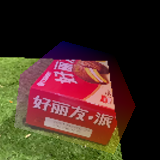} \\
    \SCRfigline{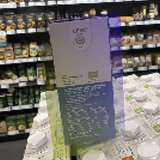} \\
    \SCRfigline{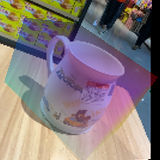} \\
    \SCRfigline{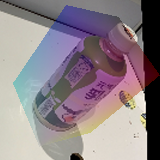} \\
    \SCRfigline{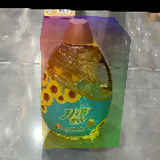} \\
    \SCRfigline{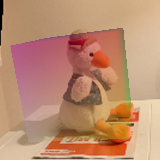} \\
    \end{SCRfigtabular}}
    \\[-0.2cm]
    \captionof{figure}{\textbf{Regression examples} on the Onepose dataset~\cite{onepose}.  Best viewed in color.}
    \label{fig:regression_results_onepose}
\end{center}
\end{figure*}

\begin{figure*}
\begin{center}
    \resizebox{\linewidth}{!}{
    \begin{SCRfigtabular}
    \SCRfigline{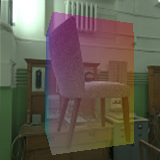} \\
    \SCRfigline{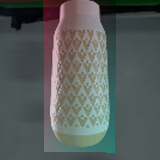} \\
    \SCRfigline{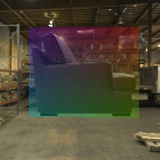} \\
    \SCRfigline{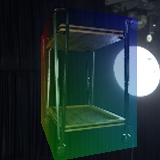} \\
    \SCRfigline{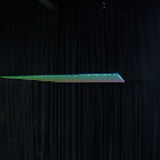} \\
    \SCRfigline{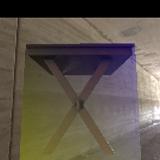} \\
    \SCRfigline{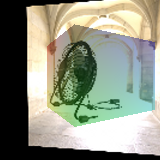} \\
    \SCRfigline{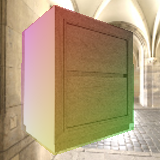} \\
    \end{SCRfigtabular}}
    \\[-0.2cm]
    \captionof{figure}{\textbf{Regression examples} on the Amazon dataset~\cite{abo_dataset}.  Best viewed in color.}
    \label{fig:amazon_examples}
\end{center}
\end{figure*}

We present qualitative results for the Linemod, OnePose and ABO datasets, in Figures~\ref{fig:regression_results_linemod2},~\ref{fig:regression_results_onepose} and \ref{fig:amazon_examples}.
Each row shows, for a given query object, a subset of the reference views of the same object overlaid with the cuboid proxy shape, as well as the predicted coordinates and confidence map. 
For visualization purposes, we only display coordinates with a confidence above a predefined threshold ($\tau=2.5$).
In the query image, we also show the ground-truth and predicted pose using green and blue boxes, respectively. 
Our approach is able to infer a plausible pose for the target object, precisely enough to meet most of the time the high accuracy standards of object-specific benchmarks. 

\begin{figure*}
\begin{center}
    \captionsetup{type=figure}
    \resizebox{\linewidth}{!}{
    \begin{SCRfigtabular}
    \SCRfigline{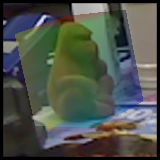} \\
    \SCRfigline{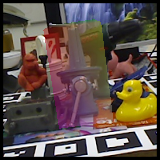} \\
    \SCRfigline{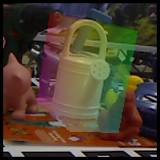} \\
    \SCRfigline{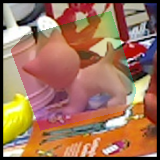} \\
    \SCRfigline{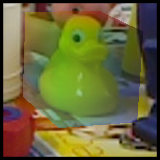} \\
    \SCRfigline{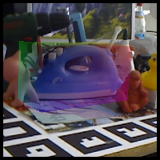} \\
    \SCRfigline{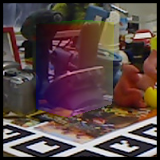} \\
    \end{SCRfigtabular}}
    \\[-0.2cm]
    \captionof{figure}{\textbf{Failure examples} on the LINEMOD dataset. Best viewed in color.}
    \label{fig:failure_results_linemod}
\end{center}
\end{figure*}

We also show examples where our method specifically fails in %
Figure~\ref{fig:failure_results_linemod}.
Failures cases typically happen with symmetric objects, or non-symmetric objects having a complex 3D shape (note how the confidence is lower in such cases). 
For more examples, we attached a anonymous video link showing predictions of our method on the whole LINEMOD and OnePose datasets. The video shows 3D bounding boxes using predicted pose and ground truth pose respectively, as well as the predicted coordinates and confidence map. The video url is as follow: \url{https://drive.google.com/file/d/1xIoyFC825487f1qFkKaUN99bEmD64OML/view?usp=sharing}

\section{Evaluation on full ABO test dataset}
\label{sec:full_abo_evaluation}

For the sake of completeness, we report performance on the official ABO test split~\cite{abo_dataset} in Table~\ref{tab:full_abo}.
Since this dataset has not been used yet for evaluation to the best of our knowledge, we propose the following evaluation protocol.
For each object in the material benchmark dataset, several environment maps are used to render the object with different lighting and background conditions
(typically, 3 environment maps per object are provided) as well as an empty map (\ie a black background).
We only use images with environment maps and discard renderings with a black background for this experiment.
We select reference views among images from the first environment and having an even index, and select query views among images using the other environments and having an odd index.
The split between even and odd indices ensures that object poses in query views are never seen in the reference images.
Although objects categories are shared between the train and test splits, our method is able to generalize to unseen object instances without any explicit input of the object category at test time.

\begin{table*}[ht!]
    \centering
    {\small
    \setlength\tabcolsep{2pt} %
    \begin{tabular}{c|cccc} 
    \toprule
      & \small median pose error~$\downarrow$ & \small 1cm-1deg~$\uparrow$ & \small 3cm-3deg~$\uparrow$ & \small 5cm-5deg~$\uparrow$ \\ 
    \midrule
    $K = 16$ & $1.4cm, 1.1^{\circ}$ & 27.61 & 64.84 & 74.47  \\
    $K = 32$ & $1.2cm, 1.0^{\circ}$ & 33.07 & 68.25& 76.17  \\
    $K = 64$ & $1.1cm, 0.9^{\circ}$ & 35.11 & 69.19 & 76.64 \\
    
    \bottomrule
    \end{tabular}
    }
    \caption{\textbf{Results on the full ABO test split} for different numbers $K$ of reference images.
    }

    \label{tab:full_abo}
\end{table*}

\section{Impact of proxy shape variability}
\label{sec:proxy_aug}

While our approach relies on a proxy shape of the object (typically a 3D bounding box), we do not need this shape to be precisely annotated, which simplifies deployment in real-life settings. To demonstrate this, we conduct an ablation where we perturb the proxy bounding box, using uniform random 3D rotations, uniform random translations up to 10\% of the original box size, and up- or down-scaling of 10\% of the bounding box size. Results in Table~\ref{tab:ab_proxy_aug} for the LINEMOD and OnePose benchmark show that the perturbations do not highly impact the performance on our method. This means that our method is robust to imprecision in the size/position/orientation of the proxy shape.

\begin{table*}[ht!]
    \centering
    {\small
    \setlength\tabcolsep{2pt} %
    \begin{tabular}{c|cc|ccc} 
    \toprule
    \multirow{2}{*}{} & \multicolumn{2}{c|}{LINEMOD} & \multicolumn{3}{c}{OnePose} \\ 
      & \small ADD(S)-0.1d~$\uparrow$ & \small Proj2D~$\uparrow$ & \small 1cm-3deg~$\uparrow$  & \small 3cm-3deg~$\uparrow$ & \small 5cm-5deg~$\uparrow$\\ 
    \midrule
    baseline & 68.4 & 90.3 & \bf 28.5 & 76.3 & 87.8\\
    random rotations & 68.2 & 89.7 & 27.8 & \bf 76.5 & \bf 87.9\\
    random rotations \& translations & 67.3 & 89.2 & 26.9 & 75.9 & 87.7 \\
    random rotations \& down-scaling & \bf 69.4 & \bf 90.7 & 26.7 & 76.0 & 87.8\\
    random rotations \& up-scaling & 68.4 & 90.3 & 26.8 & 75.3 & 87.4 \\
    random rotations \& translations \& scaling & 67.3 & 89.0 & 26.2 & 75.4 & 87.5\\

    \bottomrule
    \end{tabular}
    }
    \caption{\textbf{Ablation on the perturbation of the proxy bounding box.}
    }
    \label{tab:ab_proxy_aug}
\end{table*}

\section{Model to Model estimation}
\label{sec:m2m_eval}

To get additional insights, we experiment with a `model-to-model' prediction mode, instead of `cuboid-to-cuboid' previously.
Specifically we train our model to estimate the 3D surface's coordinates of the query object, given the 3D object surface's coordinates in the reference frames as input (instead of the proxy shape). 
We obtain the object coordinates by combining the object pose with dense depth or 3D CAD model, see right-hand side of Figure~\ref{fig:proxy_shape}. 
Results in Table~\ref{tab:ab_m2m_b2b} shows that our method performs slightly better in this setting.
This is expected, since the prediction task is now simpler: output coordinates are `attached' to the object surface, instead of floating in the air (or inside the object) as was the case for the invisible proxy shape.
While still one-shot and RGB-only at inference time, this setting is however obviously less scalable than using the proxy shape since it requires either depth maps or CAD models.

\begin{table*}[ht!]
    \centering
    \setlength\tabcolsep{3pt} %
    \resizebox{1.0\textwidth}{!}{
    \begin{tabular}{c| c c c c c c c c c c c c c |c} %
    
    \toprule %
        \multirow{2}*{Type} & \multicolumn{13}{c|}{Object Name} & \multirow{2}*{Avg.} \\
        &ape & benchwise & cam & can & cat & driller & duck & eggbox${}^{*}$ & glue${}^{*}$ & holepuncher & iron & lamp & phone & \\ 
        \hline
        \multicolumn{1}{c|}{} & \multicolumn{13}{c|}{\textit{ADD(S)-0.1d}} &  \\ \hline
        
        \textbf{Cuboid} & 47.2&      73.5&      87.5&      85.4&      80.2&      92.4&      60.8&      99.6&      69.7&      93.5&      82.4&      95.8&      51.6&     78.4\\
        \textbf{Model} & 29.6&      98.1&      88.9&      99.4&      86.6&      97.6&      45.7&      99.6&      89.4&      65.0&      97.8&      77.4&      80.4&     \bf 81.2\\
        \midrule %

        \multicolumn{1}{c|}{} & \multicolumn{13}{c|}{\textit{Proj2D}} & \\ \hline
        
        \textbf{Cuboid} &97.1&      94.1&      98.4&      98.2&      98.4&      95.7&      96.3&      99.0&      94.8&      99.3&      94.6&      94.2&      88.9&     96.1\\
        \textbf{Model} & 96.3&      98.6&      98.9&      99.0&      98.9&      97.3&      98.0&      98.9&      92.8&      94.4&      97.3&      96.3&      92.3&     \bf 96.8\\
        
        \bottomrule %
        \end{tabular}
    }
    \caption{\textbf{Ablation on the input coordinates type of Pose Encoder.} 
    The number of reference images ($K$) used is 64. 
    }
    \label{tab:ab_m2m_b2b}
\end{table*}

\section{Implementation details}
\label{sec:technical_details}

\subsection{Pose encoding}
\begin{figure}
\centering
\includegraphics[width=0.2\linewidth]{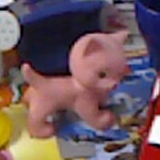}
\includegraphics[width=0.2\linewidth]{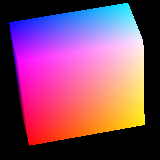}
\includegraphics[width=0.2\linewidth]{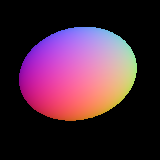}
\includegraphics[width=0.2\linewidth]{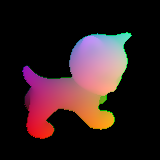}
\caption{\label{fig:proxy_shape} \textbf{Example of \emph{proxy shapes}} considered in this study. RGB image, \emph{cuboid}, \emph{ellipsoid}, and \emph{model} shapes, from left to right respectively. Reference coordinates are color-coded for visualization.}
\end{figure}

To encode information about the pose and shape of the object in a reference image, we provide the model with a pointmap featuring \emph{reference coordinates} of a \emph{proxy shape}, illustrated in Figure~\ref{fig:proxy_shape}.

\PAR{Reference coordinates}
We define for each object a reference coordinates system allowing to express its pose numerically. 
This reference coordinates system is defined based on an axis-aligned bounding box of the object, so that coordinates of points within the bounding box of the objects spread within a range of (-1,1).

\PAR{Proxy shape} 
We experiment with different proxy shapes: 
a 3D cuboid whose orientation and size matches the 3D object bounding box (\emph{cuboid}), a 3D ellipsoid whose axes matches the 3D object bounding box (\emph{ellipsoid}), and the surface of the object defined by a 3D mesh when available (\emph{model}).
In all cases, we render the proxy shape using the known camera parameters into a pointmap that associates to each pixel the \emph{reference coordinates} of the corresponding point on the \emph{proxy shape}. We arbitrarily assign null coordinates to pixels outside of the proxy shape.

\PAR{Augmentations}
To enforce the generalizability of our model, we perform several random augmentations of the pose encoding during training. We randomly translate, rotate and scale the reference coordinates system with a uniform translation range of $\pm 10\%$ of the bounding box dimensions, a uniformly sampled 3D rotation, and a uniform scaling range of $\pm 10\%$. Similarly, we randomly translate and scale the proxy shape with a uniform translation range of $\pm 10\%$ of the bounding box dimensions, and a uniform scaling range of $\pm 10\%$ when not using the \emph{model} shape.

\subsection{Memory optimization}

We now explain the memory-optimized training and inference.
In the main paper, we expose the forward pass and training details in \emph{Training details} section. 
As a reminder, we feed the network with batches of $B\times(N_{Q}+N_{R})=16\times(16+32)=768$ images, where $B$ denotes the number of unique objects per batch, $N_{Q}$ the number of queries per object and $N_{R}$ the number of reference views per object. 
The $B \times N_Q$ query features $\{\F_{b,q}\}_{b=1..B,q=1..N_Q}$ are first computed individually using the ViT image encoder (See \emph{Model architecture} Section from the main paper).
Likewise, the $B \times N_R$ reference features $\{\F_{b,i}\}_{b=1..B,i=1..N_R}$ are computed individually using the image and pose encoders. 
At the batch level, we can represent these features as two tensors %
\[
\begin{cases}
F_Q \in \mathbb{R}^{B\times N_Q \times \SQ \times D}\\
F_R \in \mathbb{R}^{B\times N_R \times \SR \times D},\\
\end{cases}
\]
where $S$ is the sequence length (number of tokens per view) and $D$ is the token dimension.

Once all features are extracted, our decoder combines information from each query with all reference views from the same object using cross-attention.
In the original transformer paper~\cite{attn}, attention is defined as a function $\fattn: (X, Y) \rightarrow X'$, where $X,X'\in\mathbb{R}^{B\times S \times D}$ and $Y\in\mathbb{R}^{B\times S' \times D}$.
The role of $\fattn$ is to make all $S$ tokens from $X$ attend all $S'$ tokens from $Y$, and this operation is performed independently over the batch dimension $B$.
Note that $\fattn$ is often defined as a function of 3 intermediate tensors $Q, K$ and $V$, but since $(Q,K,V) = f_{\text{proj}}(X,Y)$, these definitions are equivalent for all practical purposes.
We refer to $\fattn$ as ``vanilla attention'' in this section.

In our approach, as in the original transformer~\cite{attn}, the decoder is composed of a series of blocks themselves comprising 3 modules (self-attention, cross-attention, and a MLP).
We now describe in details how we can exploit the vanilla attention $\fattn$ in spite of having one extra tensor dimension in $F_Q$ and $F_R$:
\begin{enumerate}
    \item \textbf{Self-attention} is performed individually for each query view. We therefore pack all query views in the batch dimension, \ie we reshape $F_Q$ as $F'_Q\in\mathbb{R}^{(B \times N_Q) \times \SQ \times D}$ in-place and compute $\fattn(F'_Q,F'_Q)$.

    \item \textbf{Cross-attention} makes all $\SQ$ tokens from a \emph{single} query view attend all tokens from \emph{all} reference views of the same object ($N_R S$ tokens in total). 
    We achieve that by packing views on the sequence dimension, \ie we reshape $F_Q$ as $F''_Q\in\mathbb{R}^{B \times (N_Q \times \SQ) \times D}$ and $F_R$ as $F''_R\in\mathbb{R}^{B \times (N_R \times \SR) \times D}$ (again, both are in-place) and compute $\fattn(F''_Q,F''_R)$.
    Note that here, we pack all query views together, which is seemingly contradictory with our needs. However, this is equivalent to feeding query views separately since the query tokens do not interact with each other during cross-attention.
    
    \item \textbf{The MLP} processes all tokens individually. We thus pack all query views on the batch dimension as mentioned above for the self-attention case.
\end{enumerate}

\PAR{Comparison to naive implementation}.
In a naive implementation, one would need to reshape and expand $F_Q$ and $F_R$ as, respectively, $(B \times N_Q) \times S \times D$ and $(B \times N_Q) \times (N_R \times S) \times D$ tensors.
This would thus increase the memory footprint for $F_R$ by a factor $N_Q$, which corresponds to an additional 5GB memory requirement for our particular experimental settings.

\subsection{Detailed training settings}

We report in Table~\ref{tab:training_setting} the detailed parameter setting we used in our training. %

\begin{table*}[]
\begin{center}
\begin{tabular}{ll}
\toprule
Hyperparameters & Value \\
\midrule
Optimizer & AdamW~\cite{adamw} \\
Adam $\beta$ & (0.9, 0.95) \\
Learning rate scheduler &  Cosine decay \\
Training epoch & 40 \\
Warmup epochs & 4 \\
Base learning rate & 1e-4 \\
Min Learning rate & 1e-6 \\
Weight decay & 0.05 \\
Batch size & $16 \times (16 + 32) = 768$ images \\
\# of unique object per batch & 16 \\
\# of reference images per object & 32 \\
\# of query images per object & 16 \\
\midrule
Input resolution & $224 \times 224$ \\
{Background error $E$} & 1\\
\bottomrule
\end{tabular}
\caption{\textbf{Detailed training setting.}}
\label{tab:training_setting}
\end{center}
\end{table*}

\subsection{Compute resources.}
Training our MFOS model from scratch (excluding CroCo pretraining, since we use an off-the-shelf pretrained model) for 40,000 steps takes about 32 hours with 4 NVIDIA A100 GPUs.

\section{General considerations}
\label{sec:general_considerations}

\subsection{Assets used in this submission}

\begin{table*}[h]
\begin{center}
\resizebox{\linewidth}{!}{
\begin{tabular}{p{0mm}ll}
\toprule
\multicolumn{2}{l}{Asset} & License \\
\midrule
\multicolumn{3}{l}{\textbf{BOP dataset}~\cite{bop_dataset}} \\

& LM(Linemod)~\cite{linemod} &  Creative Commons Attribution 4.0 International (CC BY 4.0) \\
& T-LESS & Creative Commons Attribution 4.0 International (CC BY 4.0) \\
& HB (HomebrewedDB) &  Creative Commons (CC0 1.0 Universal) \\
& Hope (NVIDIA Household Objects for Pose Estimation)  &  Creative Commons Attribution-NonCommercial-ShareAlike 4.0  (CC BY-NC-SA 4.0)\\
& YCB-V (YCB-Video) &  MIT \\
& RU-APC (Rutgers APC)  &  Unknown\\
& TUD-L (TUD Light)   &  Creative Commons Attribution-ShareAlike 4.0  (CC BY-SA 4.0) \\

& TYO-L (Toyota Light)   &  Creative Commons Attribution-NonCommercial 4.0  (CC BY-NC 4.0) \\
& IC-MI (Tejani et al.)   &  Unknown \\

\midrule
\multicolumn{3}{l}{\textbf{OnePose Datasets}} \\
& OnePose~\cite{onepose} & Apache License 2.0 \\
& OnePose-LowTexture~\cite{onepose_plusplus} & Apache License 2.0 \\

\midrule
\multicolumn{3}{l}{\textbf{Amazon Berkeley Objects (ABO) Dataset}} \\
& ABO~\cite{abo_dataset} &Creative Commons Attribution-NonCommercial 4.0 (CC BY-NC 4.0)  \\

\bottomrule
\end{tabular}
}
\caption{\textbf{Licenses of assets used in our experiments.}}
\label{tab:assets_licenses}

\end{center}
\end{table*}

We provide an overview of assets used in our experiments and their licenses in Table~\ref{tab:assets_licenses}.

\subsection{Limitations of the proposed approach}
The proposed approach suffers from several limitations.
While improving from a practical and scalability point of view over existing methods that require a full 3D model of the object, the proposed approach still requires a set of reference images with pose annotations. Its implementation is currently limited to rigid non-symmetrical objects. Furthermore, it requires image crops roughly centered on the object, and thus relies on the use of a 2D object detector.

\subsection{Ethics}
This research contributes to the development of object pose estimation, with potential applications in robotics, augmented reality (AR), and machine vision in general. While many of those applications could bring societal benefits (e.g. workload decrease through automation, AR-based teaching or assistance), it could also be used for unethical purposes.

\FloatBarrier{}

\end{document}